\documentclass[10pt,twocolumn,letterpaper]{article}

\usepackage{cvpr}
\usepackage{times}
\usepackage{epsfig}
\usepackage{graphicx}
\usepackage{amsmath}
\usepackage{amssymb}
\usepackage{enumitem}
\usepackage{subfigure}
\usepackage{booktabs}
\usepackage{array}
\usepackage{gensymb}
\usepackage{multirow}

\usepackage[pagebackref=true,breaklinks=true,letterpaper=true,colorlinks,bookmarks=false]{hyperref}

\cvprfinalcopy 

\def\cvprPaperID{****} 

\ifcvprfinal\fi
\begin{document}

\title{High-Resolution Multispectral Dataset for Semantic Segmentation}

\author{Ronald Kemker, Carl Salvaggio, and Christopher Kanan\\
Chester F. Carlson Center for Imaging Science\\
Rochester Institute of Technology\\
{\tt\small \{rmk6217, cnspci, kanan\}@rit.edu}}

\maketitle

\begin{abstract}
Unmanned aircraft have decreased the cost required to collect remote sensing imagery, which has enabled researchers to collect high-spatial resolution data from multiple sensor modalities more frequently and easily.  The increase in data will push the need for semantic segmentation frameworks that are able to classify non-RGB imagery, but this type of algorithmic development requires an increase in publicly available benchmark datasets with class labels.  In this paper, we introduce a high-resolution multispectral dataset with image labels.  This new benchmark dataset has been pre-split into training/testing folds in order to standardize evaluation and continue to push state-of-the-art classification frameworks for non-RGB imagery.  
\end{abstract}

\section{Introduction}
The semantic segmentation of remote sensing imagery provides the end user a pixel-wise classification map for a given scene.  Countless machine- and deep-learning algorithms have been developed to perform this task; however, access to large quantities of labeled data for non-RGB sensors make the deployability of these frameworks difficult.  In computer vision literature, semantic segmentation has made significant progress due to deep-convolutional neural networks (DCNNs) \cite{SimonyanZ14a,he2016deep} trained with large quantities of labeled imagery \cite{ILSVRC15}.  The quantity of labeled data for multispectral (MSI) and hyperspectral imagery (HSI) is minuscule in comparison, which has made DCNNs less successful for remote sensing applications.

Common benchmark datasets were acquired by airborne and satellite platforms, so the ground-sample distance (GSD) is normally on the order of 1-20 meters.  Many of these benchmarks consist of a single image that is randomly sampled for training/testing folds. This is a problem since deployable frameworks will not be able to build training/testing folds on the fly because images captured in real-time do not have labels.  In addition, many classification frameworks involve extracting spatial information from neighboring pixels.  If the folds are randomly sampled, the training data could  contaminate the test data, which would artificially inflate classification performance~\cite{shen2011three, mirzapour2015improving, zhao2015combining}.  This is why it is important to separate training and testing data.

\begin{figure}[t]
  \centering
  \subfigure[Train]{%
  \includegraphics[width=0.25\linewidth]{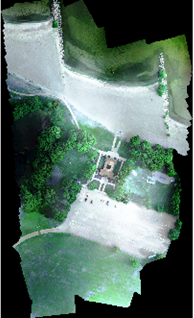}
  \label{fig:train_ortho}}
  \subfigure[Validation]{%
  \includegraphics[width=0.3225\linewidth]{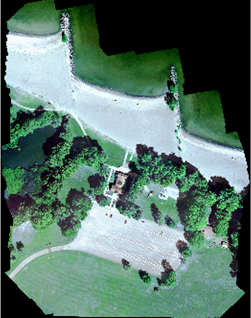}
  \label{fig:val_ortho}}
  \subfigure[Test]{%
  \includegraphics[width=0.253\linewidth]{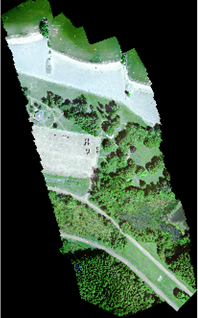}
  \label{fig:test_ortho}}
  \caption{RGB visualization of Hamlin Beach State Park dataset.}
  \label{fig:ortho}
\end{figure}

In this paper, we introduce a high-resolution (4.7cm GSD) multispectral dataset acquired by an unmanned aircraft system (UAS).  It contains 18 unbalanced classes and will be used to evaluate semantic segmentation frameworks designed for non-RGB remote sensing imagery.  This dataset, shown in Figure \ref{fig:ortho}, is split into training, validation, and testing folds to 1) provide a standard for state-of-the-art comparison, and 2) demonstrate the feasibility of deploying algorithms in a real-time environment.  Preliminary results demonstrate that the large spatial variability commonly associated with high-resolution imagery, large sample (pixel) size, small and hidden objects, and unbalanced class distribution make this a difficult dataset to perform well on; and in the future, should be a ripe candidate for deep learning frameworks.

\section{Related Work}

\subsection{Non-RGB Labeled Datasets}

The lack of publicly available labeled remote sensing imagery from non-RGB sensors has prevented the successful incorporation of DCNN-type architectures popular in the computer-vision community.  Image classification frameworks, such as VGG-16 \cite{SimonyanZ14a} and ResNet \cite{he2016deep}, are trained on the massive ImageNet dataset\cite{ILSVRC15}.  The ImageNet challenge has over a million training images for 1,000 class-types.  State-of-the-art semantic segmentation frameworks transfer the weights from DCNNs trained on ImageNet and fine-tune them on a much smaller semantic segmentation dataset like PASCAL VOC \cite{Everingham15} or MS COCO\cite{LinMBHPRDZ14}.

Labeled remote sensing datasets captured by non-RGB sensors (i.e. MSI/HSI) are much smaller than ImageNet, making this approach currently infeasible.  Instead, researchers have embraced unsupervised feature extraction as a method for improving classification performance.  These features are fed into a classifier such as a support vector machine (SVM) or multi-layer perceptron (MLP) to generate a prediction.

Publicly-available datasets with labels have traditionally been imaged by airborne and satellite platforms.  Consequently, the GSD of non-RGB imagery is on the order of 1-20 meters.  Indian Pines was one of the first publicly-available HSI\cite{PURR1947}.  It was captured by the AVIRIS airborne sensor and it has a GSD of 20 meters.  Additional benchmark HSI datasets, listed in Table \ref{table:hsi}, have been released, but none of these datasets have a GSD less than a meter.  There are also several MSI datasets publicly available, including (but not limited to) semantic segmentation challenges hosted by the International Society for Photogrammetry and Remote Sensing (ISPRS) \cite{meidow2014theme},  the 2016 IEEE Geoscience and Remote Sensing Society (GRSS) data fusion contest\cite{grss2017data}, and the Satellite Imagery Feature Detection challenge on Kaggle\cite{kaggle2017}.

\begin{table}[ht!]
\centering
\begin{tabular}{@{}llcc@{}}\toprule
\textbf{Dataset} & \textbf{Sensor} &\textbf{GSD} & \textbf{Classes} \\ \midrule
\multirow{ 2}{*}{GRSS Data Fusion '16}& Landsat/&\multirow{ 2}{*}{100} & \multirow{ 2}{*}{17}\\
& Sentinel 2& & \\
Botswana & Hyperion & 30 & 14 \\
Indian Pines & AVIRIS & 20 & 16 \\
Kennedy Space Center & AVIRIS & 18 & 13 \\
Salinas Valley & AVIRIS & 3.7 & 16 \\
Pavia University & ROSIS & 1.3 & 9\\
Pavia Center & ROSIS & 1.3 & 9 \\
Kaggle Challenge & World-View 3 & 0.3-7.5 & 10\\
ISPRS Vaihingen & 4-band MSI & 0.09 & 6 \\
ISPRS Potsdam & 4-band MSI & 0.05 & 6 \\
\textbf{Our Dataset} & \textbf{6-band MSI} & \textbf{0.047} & \textbf{18} \\
\bottomrule \\
\end{tabular}
\caption{Benchmark semantic segmentation datasets for non-RGB imagery, the sensor that collected it, its ground sample distance (GSD) in meters, and the number of classes.}
\label{table:hsi}
\end{table}

UAS collection of non-RGB imagery has grown in popularity, especially in precision agriculture, because it is more cost effective than manned flights and provides better spatial resolution than satellite imagery.  This cost savings allows the user to collect data more frequently, which increases the temporal resolution of the data as well.  The authors in \cite{huang2010multispectral} characterized three MSI payloads on numerous applications including crop health sensing, variable-rate application prescription, irrigation engineering, and crop-field variability.  The same sensor used to build the dataset presented in this paper has also been used on-board UASs to assess crop stress by measuring the variability in chlorophyll fluorescence \cite{ZarcoTejada20091262} and through the acquisition of other biophysical parameters\cite{berni2009thermal}.  The sensor used here has also been used to perform vegetation classification on orthomosaic imagery\cite{laliberte2011multispectral}.  The authors used multi-scale segmentation and hand-selected features to identify 8 types of plant life known to be present in the scene.

In 2016, the Chester F. Carlson Center for Imaging Science established a new UAS laboratory to collect remote sensing data for research purposes.  This laboratory is equipped with several UAS payloads including RGB cameras,  MSI/HSI sensors, thermal imaging systems, and light detection and ranging (LIDAR).  In this paper, we present the first labeled dataset created by this lab, which contains 4.7 cm resolution MSI (6-band) and 18 classes.

\subsection{Separate Training/Testing Folds}

The majority of published work involving the classification of non-RGB remote sensing imagery involves the use of small, single-image datasets such as the HSI datasets listed in Table \ref{table:hsi}.  The training/testing sets are usually built by randomly sampling a percentage of the image.  Many of these papers use different training/testing sets rather than established benchmarks, which makes it difficult to 1) identify the current state-of-the-art, and 2) provide a fair comparison against other published algorithms.

The construction of training/testing folds from a single image may be useful for prototyping algorithms, but it is not representative of a deployable framework.  A pre-trained machine learning model will not have access to new labels in a deployed environment, so the model must be able to adapt to a wide range of circumstances to make generalized predictions based on data it has already seen.  To demonstrate that a model can do this, the training/testing data must be kept separate - a cardinal rule for any machine learning framework.  A good remote sensing dataset should collect training/testing folds from completely separate scenes, but it should share some of the same class labels.

Many benchmark datasets in computer vision, such as ImageNet and PASCAL VOC, encourage the development of deployable models by protecting the testing labels.  The participant is required to submit their predictions to an evaluation server in order to obtain their results.  The remote sensing community has slowly begun to establish their own benchmark evaluation servers such as the IEEE GRSS Data Fusion Contest, ISPRS semantic segmentation challenges, and more recently, new contests posted on Kaggle.  These challenges will continue to push algorithm development by standardizing evaluation and clearly identifying the state-of-the-art performer.  The dataset introduced in this paper has a separate training, validation, and testing fold, and we are working with the IEEE GRSS to make it available on their evaluation server.

\section{Data Collection}

\subsection{Collection Site}

The imagery for this dataset was collected at Hamlin Beach State Park, located along the coast of Lake Ontario in Hamlin, NY.  The training and validation data was collected at one location, and the test data was collected at a different location in the park.  These two locations are unique, but they share many of the same class-types.  Table \ref{table:collection} lists several other collection parameters that may be of interest.

\begin{table}[ht!]
\begin{tabular}{@{}lccc@{}}\toprule
 & \textbf{Train} & \textbf{Validation} & \textbf{Test} \\ \midrule 
 \textbf{Date}&29 Aug 16&6 Sep 16 &29 Aug 16\\
 \textbf{Time (UTC)}&13:37 &15:18 &14:49\\
 \textbf{Weather} & \multicolumn{3}{c}{Sunny, clear skies}\\
 \textbf{Solar Azimuth}& 109.65\degree
 & 138.38\degree &126.91\degree\\
 \textbf{Solar Elevation}& 32.12\degree & 45.37\degree &43.62\degree \\
\bottomrule \\
\end{tabular}
\caption{Collection parameters for training, validation, and testing folds for dataset.}
\label{table:collection}
\end{table}

\subsection{Collection Equipment}

\begin{figure}[ht!]
\begin{center}
   \includegraphics[width=0.7\linewidth]{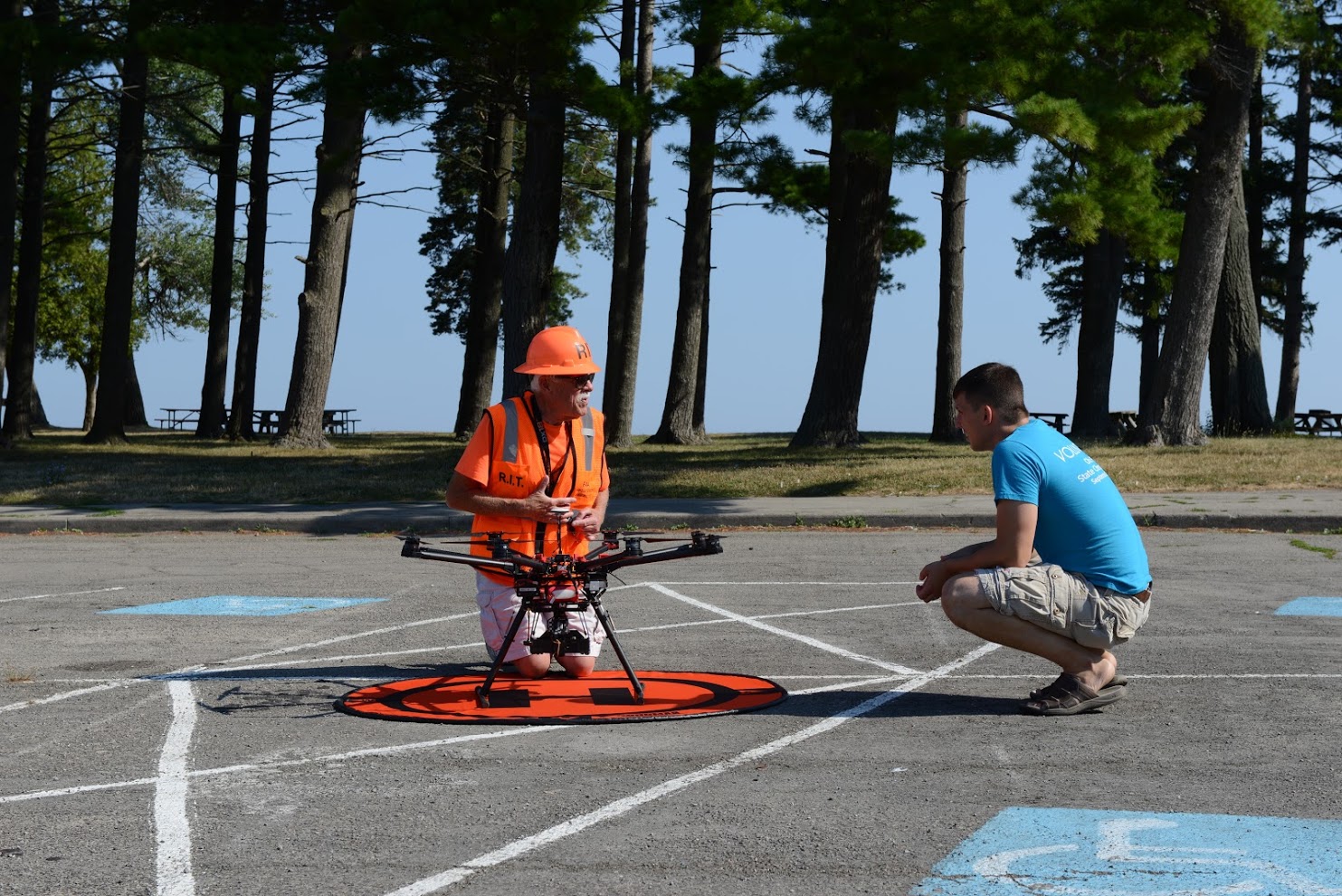}
\end{center}
   \caption{Tetracam Micro-MCA6 mounted on-board the DJI-S1000 octocopter prior to collection.}
\label{fig:octocopter}
\end{figure}

The equipment used to build this dataset and information about the flight is listed in Table \ref{table:specs}.  The Tetracam Micro-MCA6 MSI sensor has six independent optical systems with bandpass filters centered across the visual and near-infrared (VNIR) spectrum.  Figure \ref{fig:octocopter} shows an image of the Micro-MCA6 mounted on-board the DJI-S1000 octocopter. 

\begin{table}[ht!]
\centering
\begin{tabular}{@{}ll@{}}\toprule
\multicolumn{2}{l}{\textbf{Imaging System}} \\
\midrule 
Manufacturer/Model & Tetracam Micro-MCA6\\
Spectral Range [nm] & 490-900\\
Spectral Bands & 6 \\
RGB Band Centers [nm] & 490/550/680 \\
NIR Band Centers [nm] & 720/800/900 \\
Spectral FWHM [nm] & 10 (Bands 1-5)\\
                   & 20 (Band 6)\\
Sensor Form Factor [pix] & 1280x1024 \\
Pixel Pitch [$\mu$m] &  5.2 \\
Focal Length [mm]& 9.6\\
Bit Resolution & 10-bit \\
Shutter & Global Shutter \\
\midrule
\multicolumn{2}{l}{\textbf{Flight}} \\ \midrule
Elevation [m] & 120 (AGL) \\
Speed [m/s] & 5 \\
Ground Field of View [m] & $\approx$60x48\\
GSD [cm] & 4.7\\
Collection Rate [images/sec] & 1 \\
\bottomrule \\
\end{tabular}
\caption{Data Collection Specifications}
\label{table:specs}
\end{table}

\section{Data Processing}

\subsection{Pre-Processing}
\label{section:preprocessing}
\begin{figure}[ht]
  \centering
  \subfigure[Affine]{%
  \includegraphics[width=0.445\linewidth]{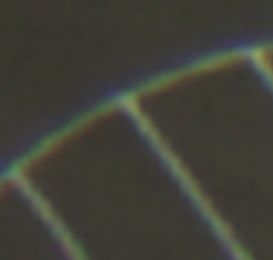}
  \label{fig:affinecal}}
  \subfigure[Perspective]{%
  \includegraphics[width=0.4655\linewidth]{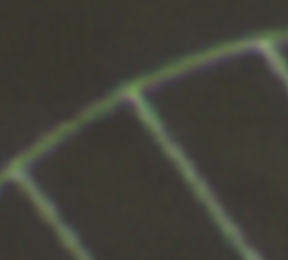}
  \label{fig:perspectivecal}}
  \caption{Difference between manufacturer's affine transformation, Figure \ref{fig:affinecal}, and our perspective transformation, Figure \ref{fig:perspectivecal}.  The registration error in the affine transformation looks like a blue and red streak along the top and bottom of the parking lines, respectively.}
  \label{fig:transformation}
\end{figure}

For each collection campaign, we filtered out data not collected along our desired flight path (\emph{i.e.} takeoff and landing legs).  The six spectral images come from independent imaging systems, so they need to be registered to one another.  The manufacturer provided an affine transformation matrix that was not designed to work at the flying height this data was collected at which caused noticeable registration error.  We used one of the parking lot images to develop a global perspective transformation for the other images in our dataset.  Figure \ref{fig:affinecal} illustrates the registration error caused by the affine transformation provided by the manufacturer, and Figure \ref{fig:perspectivecal} shows that this error has been reduced with our perspective transformation.

The global transformation worked well for some of the images, but there were misregistration errors in other parts of the scene indicating that the transformation needs to be performed on a per-image basis.  This was done by matching SIFT features from each band to build custom homographies.  If a good homography could not be found, the global transformation was used instead.  This was common for homogenous scenes elements such as water or repeating patterns such as an empty parking lots.

Each collected frame was acquired with a unique integration time (\emph{i.e.} auto exposure) and each band of the Tetracam Micro-MCA6 uses a different integration time proportional to the sensor's relative spectral response.  Another issue is that each image, including each band, is collected with a different integration time.  The longer integration times required for darker images, especially over water scenes, resulted in blur caused by platform motion.  We normalized each image with its corresponding integration time and then contrast-stretched the image back to a 16-bit integer using the global min/max of the entire dataset.  The original images are 10-bit, but the large variation in integration time groups most of the data to lower intensity ranges.  We extended the dynamic range of the orthomosaic by stretching the possible quantized intensity states.

\subsection{Orthomosaics}
\label{section:ortho}

Agisoft PhotoScan \cite{photoscan2016} was used to build the orthomosaics from the individual images in Section \ref{section:preprocessing}.  The PhotoScan workflow involves:

\begin{enumerate}
\setlength\itemsep{0em}
\item Find key points in the images and match them together as tie-points.
\item Build a dense point cloud from the image data.
\item Build a 3D mesh and corresponding UV texture map from the dense point cloud.
\item Generate an orthomosaic onto the WGS-84 coordinate system using the mesh and image data.
\item Manually correct troublesome areas by removing photographs caused by motion blur or moving objects.
\end{enumerate}

PhotoScan can generate high-quality orthomosaics, but manual steps were taken to ensure the best quality.  First, not all of the images were in focus; and although PhotoScan has an image quality algorithm, we opted to manually scan and remove the defocused images.  Second, the 3D model that the orthomosaic is projected onto is built from structure-from-motion.  Large objects that move over time, such as tree branches blowing in the wind, or vehicles moving throughout the scene, will cause noticeable errors.  This is corrected by highlighting the affected region and manually selecting a single (or a few) alternative images that will be used to generate that part of the orthomosaic, as opposed to those automatically selected.

\section{Proposed Dataset}
\label{section:statistics}

\subsection{Training/Testing Split}

The Hamlin Beach State Park dataset (Figure \ref{fig:ortho}) is split up into a training, validation, and testing fold. Each fold contains an orthomosaic image and corresponding classification map.  Each orthomosaic contains the six-band image described in Section \ref{section:ortho} along with a mask where the image data is valid.  The spatial dimensionality for each fold is 9,393$\times$5,642 (train), 8,833$\times$6,918 (validation), and 12,446$\times$7,654 (test).

\subsection{Classification Labels}

Table \ref{table:classlabels} lists the 18 class labels for this dataset.  Each orthomosaic was hand-annotated using the region-of-interest (ROI) tool in ENVI.  Several individuals took part in the labeling process. 

\begin{table}[ht!]
\begin{tabular}{@{}ll|ll@{}}\toprule
1. & Road Markings & 10. & Orange Landing Pad\\
2. &Tree& 11. &Buoy\\
3. &Building&12. &Rocks\\
4. &Vehicle &13. &Low-Level Vegetation\\
5. &Person&14. &Grass/Lawn\\
6. &Lifeguard Chair&15. &Sand/Beach\\
7. &Picnic Table&16. &Water (Lake)\\
8. &Black Wood Panel&17. &Water (Pond)\\
9. &White Wood Panel&18. &Asphalt\\
\bottomrule \\
\end{tabular}
\caption{Class labels for the Hamlin Beach State Park dataset.}
\label{table:classlabels}
\end{table}

\begin{figure}[ht!]
\begin{center}
   \includegraphics[width=0.99\linewidth]{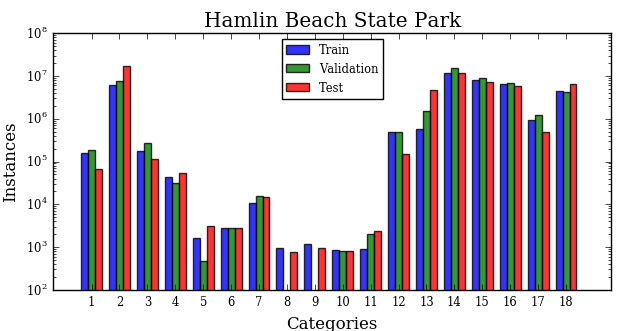}
\end{center}
   \caption{Class-label instances for the Hamlin Beach State Forest dataset.  Note: The y-axis is logarithmic to account for the number disparity among labels.}
\label{fig:stats}
\end{figure}

The class-labeled instances are, as illustrated in Figure \ref{fig:stats}, orders of magnitude different from one-another.  These underrepresented classes should make this dataset more challenging.  

\subsubsection{Water/Beach Area}
The two classes for water are lake and pond.  The lake class is for Lake Ontario, which is north of the beach.  The pond water class is for the small inland pond, present in all three folds, which is surrounded by marsh and trees.  Along Lake Ontario is a sand/beach class.  This class also includes any spot where sand blew up along the asphalt walking paths.  Along the beach are some white-painted, wooden lifeguard chairs.  The buoy class is for the water buoys present in the water and on the beach.  They are very small, primarily red and/or white, and assume various shapes.  The rocks class is for the large breakwater along the beach.

\subsubsection{Vegetation}
There are three vegetation classes including grass, trees, and low-level vegetation.  The tree class includes a variety of trees present in the scene.  The grass includes all pixels on the lawn.  There are some mixtures present in the grass (such as sand, dirt, or various weeds), so the classification algorithm will need to take neighboring pixel information into account.  The grass spots on the beach and asphalt were labeled automatically using a normalized-difference vegetation index (NDVI) metric.  The low-level vegetation class includes any other vegetation, including manicured plants, around the building or the marsh next to the pond. 

\subsubsection{Roadway}
The asphalt class includes all parking lots, roads, and walk-ways made from asphalt, but it does not include cement or stone paths around the buildings.  The road marking class is for any painted asphalt surface including parking/road lanes.  This class was automatically labeled with posteriori manual clean-up.  The road markings in the validation image are sharper than those depicted in the training image since the park repainted the lines between collects.  The vehicle label includes any car, truck, or bus.

\subsubsection{Underrepresented Classes}
Underrepresented classes, which may be small and/or appear infrequently, will be difficult to identify.  Since some of the land cover classes are massive in comparison, the mean-class accuracy metric will be the most important during the classification experiments in Section \ref{section:classification}.  Small object classes, such as person and picnic table, represent only a minute fraction of the image and should remain very difficult to correctly classify.  These small objects will be surrounded by larger classes and may even hide in the shade.

There are also a few classes that are only present in the scene a couple of times, such as the white/black wood targets, orange UAS landing pad, lifeguard chair, and buildings.  The building class is primarily roof/shingles of a few buildings found throughout the scene.  The similarity between the white wooden reflectance calibration target and the lifeguard chair should make semantic information in the scene vital to classification accuracy.  There is only a single instance of the orange UAS landing pad in every fold.  The black and white targets are not present in the validation fold, which could make it difficult to cross-validate for a model that can correctly identify them.

\section{Benchmark Results}

\subsection{Semantic Segmentation}
\label{section:classification}

The main goal of this dataset is to push the state-of-the-art for semantic segmentation of non-RGB imagery.  This section will provide some benchmark results for future development.  The training and validation folds are used to cross-validate for hyperparameters, and then the model is fit with the two folds combined using these hyperparameters.  The test data and labels are never used to cross-validate for hyperparameters or fit the model.

\subsubsection{Spectral-Only Features}
\label{section:spectral}
This paper will explore three spectral-only classification methods including $k$-nearest neighbor (kNN), linear SVM and MLP.  Spectral-only classification does not take neighboring pixel information into account, so the high spatial variability commonplace in small GSD imagery adversely effects classification performance.  Additionally, semantic information for objects present in small GSD imagery require frameworks that gather pixels from a wider receptive field.

We chose the LIBLINEAR \cite{fan2008liblinear} implementation of SVM for its speed and stability with large datasets.  This implementation uses L2-loss and a one-vs-rest multi-class approach.  We also attempted to use the radial basis function (RBF) kernel for SVM, but the large number of samples prevented the classifier from fitting properly.  The only pre-processing step was standard normalization (channel-mean subtraction and dividing by each channel's standard deviation).  We used the training and validation set to cross-validate for the cost parameter, and then performed a final fit with the combined training and validation set.

The MLP is a fully-connected neural network with a single hidden-layer.  This hidden layer is preceded by a batch-normalization layer and followed by a ReLU activation.  Since the distribution of class-labels is uneven, we assigned the class weights, $w_i$, in Equation \ref{eq:weight} to each class $i$, where $N$ is the number of samples, $N_i$ is the number of samples in each class, and $\mu$ is a tunable hyper-parameter.  The MLP is also helpful with spatial-spectral feature extraction methods in Section \ref{section:spatial} where the dimensionality is increased, making the SVM solution unstable.

\begin{equation}
\label{eq:weight}
w_i = \mu \log\left( \frac{N}{N_i} \right)
\end{equation}

In addition, we explored the impact of the additional NIR spectral bands on spectral-only classification performance.  This includes only the RGB bands (SVM-RGB), only the three NIR bands (SVM-NIR), a false-color image (SVM-CIR), and a four band RGB-NIR (SVM-VNIR).  The 720 nm band was used for the SVM-CIR and SVM-VNIR experiments.

\subsubsection{Spatial-Spectral Features}
\label{section:spatial}
The resolution of this dataset encourages the use of neighboring pixel information to improve upon spectral-only classification performance.  This paper will explore multiple spatial-spectral feature extraction techniques including mean-pooling (MP), multi-scale independent component analysis (MICA), and stacked convolutional autoencoders (SCAE).  The MP method reduces some of the spatial variability in the scene, yielding a slightly better result.  The mean-pooled response is fed into the same SVM outlined in Section \ref{section:spectral}.

\begin{figure}[ht!]
\begin{center}
   \includegraphics[width=0.75\linewidth]{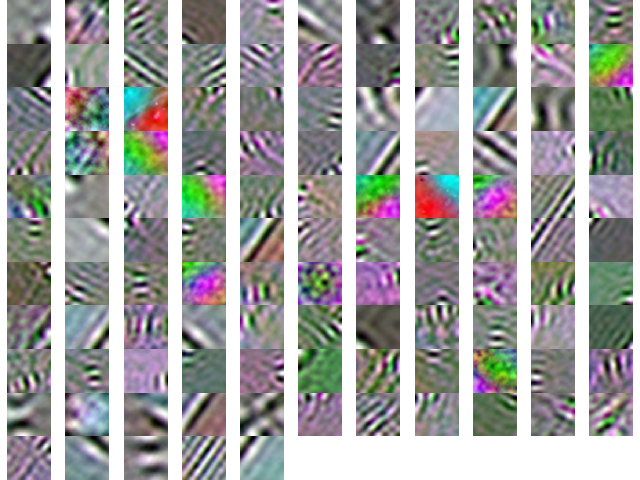}
\end{center}
   \caption{RGB visualization of MICA filters built from Tetracam Micro-MCA6 data.}
\label{fig:filters}
\end{figure}

MICA is an unsupervised low-level spatial-spectral feature extractor that learns a set of Gabor-type bar/edge and color-opponency detectors from natural images~\cite{kemker2017self}.  These filters are built by a extracting $N$ patches from the training data using a $C\times C$ patch size, vectorizing each patch to form a $N\times6C^2$ array, and then passing the patch array through a non-linear activation.  Whitened principal-component analysis (WPCA) is used to reduce the dimensionality of the patch array to $N \times F$, where $F$ is the number of desired MICA filters.  Independent component analysis (ICA) is used to break the whitened patch array into its statistically independent components.  The filters, shown in Figure \ref{fig:filters}, were constructed by multiplying the ICA and PCA vectors and then reshaping this into a $F\times C\times C\times 6$ array.   

These learned filters, as part of a shallow neural network, are convolved with the labeled imagery, passed through an activation function to introduce non-linearities, and then pooled to incorporate translation invariance.  These feature responses can then be fed to a traditional classifier. Full details are available in \cite{kemker2017self}.

\begin{figure}[ht!]
\begin{center}
   \includegraphics[width=0.95\linewidth]{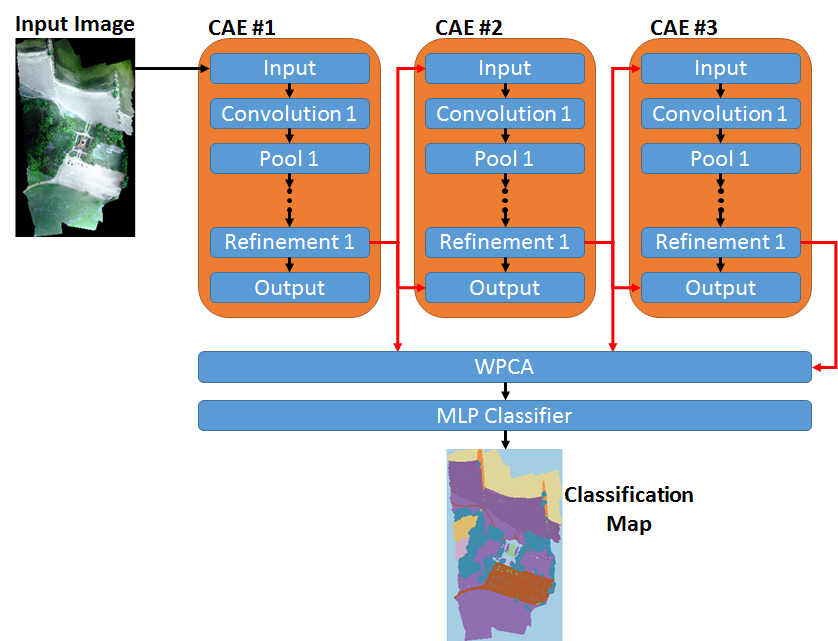}
\end{center}
   \caption{SCAE model used in this paper.  Architecture details for each CAE included in Figure \ref{fig:scae_parts}.}
\label{fig:scae}
\end{figure}

SCAE, illustrated in Figure \ref{fig:scae}, is another unsupervised spatial-spectral feature extractor.  SCAE has a deeper neural network architecture than MICA and is capable of extracting higher-level features \cite{kemker2017self}.  The architecture used in this paper involves three individual convolutional autoencoders (CAEs) that are trained independently.  The input and output of the first CAE is a collection of random image patches from the training data, and the input/output of the subsequent CAEs are the features from the last hidden layer of the previous CAE.  The output of all three CAEs are concatenated in the feature domain, mean-pooled, with the dimensionality reduced to 99\% of the original variance using WPCA.  The final feature response is passed to a traditional classifier.

\begin{figure*}[ht!]
\begin{center}
   \includegraphics[width=0.70\linewidth]{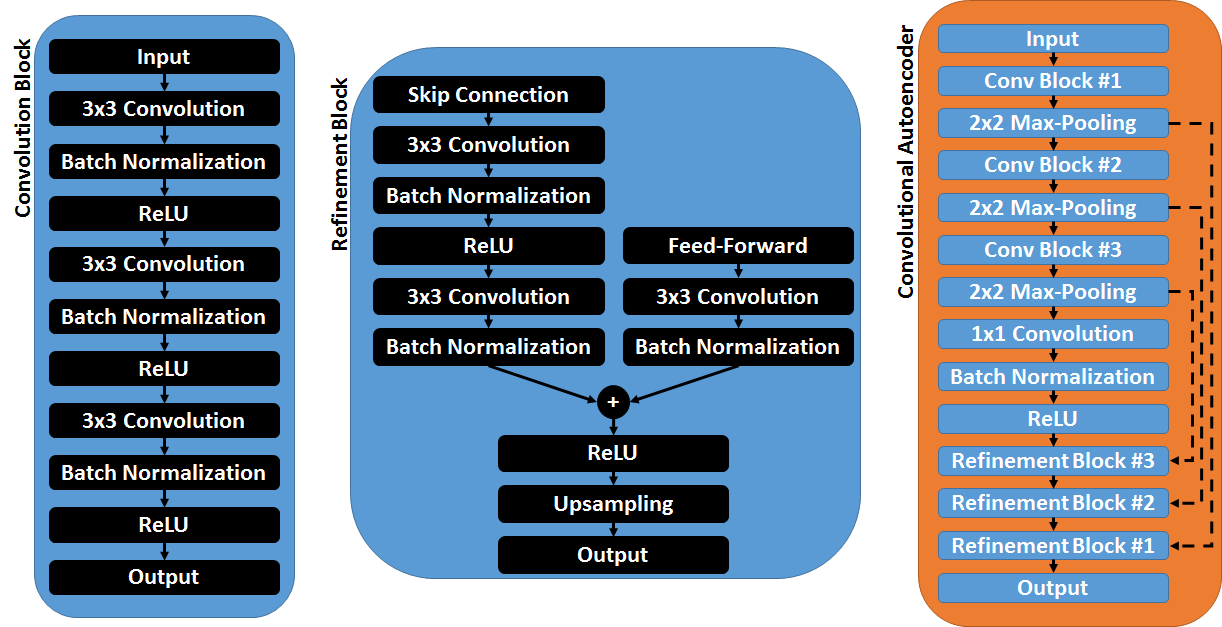}
\end{center}
   \caption{The convolutional autoencoder (CAE) architecture used in SCAE.  This CAE us made up of several convolution and refinement blocks.  The SCAE model in this paper uses three CAEs.}
\label{fig:scae_parts}
\end{figure*}

The architecture of each CAE is illustrated in Figure \ref{fig:scae_parts}.  Each CAE contains a small feed-forward network consisting of multiple convolution and max-pooling operations. The feature response is reconstructed with symmetric convolution and upsampling operations.  The reconstruction error is reduced by using skip connections from the feed-forward network, inspired by \cite{pinheiro2016learning}.

\subsubsection{Experimental Results}
\label{section:results}

The semantic segmentation results for this dataset are listed in Table \ref{table:classification}.  Each algorithm is evaluated on per-class accuracy, overall accuracy (OA), mean-class accuracy (AA), and kappa statistic ($\kappa$).  Mean-class accuracy is the most important metric for evaluating the discriminative power of our classification models due to the disparity in numbers for members in each labeled class.

\begin{table*}[ht!]
\centering
\setlength{\tabcolsep}{.25em}
\begin{tabular}{@{}lccccccccccccccccccccc@{}}\toprule
& \textbf{1} & \textbf{2} & \textbf{3} & \textbf{4}
& \textbf{5} & \textbf{6} & \textbf{7} & \textbf{8}
& \textbf{9} & \textbf{10} & \textbf{11} & \textbf{12}
& \textbf{13} & \textbf{14} & \textbf{15} & \textbf{16}
& \textbf{17} & \textbf{18} & \textbf{OA} & \textbf{AA} 
& \textbf{$\kappa$} \\ \midrule
\textbf{kNN} & 65.1 & 71.0 & 0.3 & 15.8 & 0.1 & 1.0 & 0.6 & 0.0 & 0.1 & 14.6 & 3.6 & 34.0 & 2.3 & 79.2 & 56.1 & 83.6 & 0.0 & 80.0 & 66.1 & 27.7 & 0.576\\
\textbf{SVM } & 51.0 & 43.5 & 1.5 &0.2 &19.9 &22.9 &\textbf{0.8} & \textbf{48.3} &0.3 &15.2 &0.7 &20.8 &0.4 & 71.0 & 89.5 &94.3 & 0.0 & 82.7 & 61.3 &29.6 &0.538 \\
\textbf{MLP}  & \textbf{75.6}  & 62.1  & 3.7  & 1.0  & 0.0  & 3.1
  & 0.0  & 0.0  & 0.0  & 77.4  & 1.8  & 38.8  & 0.3  & \textbf{85.4}  & 36.4  & 92.6  & 0.0  & 93.1 & 64.3 & 30.4 & 0.554\\
\textbf{MP}  & 29.6 & 44.1 & 0.6 & 0.2 & \textbf{31.2} & 16.9 & 0.6 & 47.9 & \textbf{0.8} & 22.1 & \textbf{10.1} & 33.4 & 0.1 & 73.1 & \textbf{95.2} & 94.6 & 0.2 & \textbf{93.3} & 64.0 & 31.3 & 0.568\\
\textbf{MICA} & 43.2 & \textbf{92.6} & 1.0 & \textbf{47.5} & 0.0 & \textbf{50.8} & 0.0 & 0.0 & 0.0 & 66.3 & 0.0 & \textbf{66.5} & \textbf{13.3} & 84.8 & 78.2 & 89.1 & \textbf{3.4} & 46.9 & \textbf{74.9} & \textbf{36.2} &\textbf{ 0.683}  \\
\textbf{SCAE} & 37.0 & 62.0 & \textbf{11.1 }& 11.8 & 0.0 & 29.4 & 0.0 & 0.0 & 0.4 & \textbf{82.6} & 7.2 & 36.0 & 1.1 & 84.7 & 85.3 & \textbf{97.5} & 0.0 & 59.8 & 67.4 & 32.1 & 0.594 \\ 
\bottomrule \\
\end{tabular}
\caption{Benchmark classification results for the Hamlin Beach State Park test set.  These results include per-class accuracy, overall accuracy (OA), mean-class accuracy (AA), and kappa statistic ($\kappa$).}
\label{table:classification}
\end{table*}

The kNN classifier used the Euclidian distance metric and cross-validated for $k$ over the range of 1-15.  The Linear SVM cross-validated for its cost parameter $C$ over the range of $2^{-9} - 2^{16}$ and was weighted by the inverse class frequency.  The MLP has a single-hidden layer with 64 units and uses a L2 regularization (weight-decay) value of $10^{-4}$ in the convolution and batch normalization layers.  The MLP was trained using the NAdam optimizer \cite{KingmaB14} with a batch size of 256 and the class weighted update in Equation \ref{eq:weight} where $\mu=0.15$.  The MP experiment used a $5\times 5$ filter and the feature response was passed to the same SVM described earlier.

The MICA model used in this paper had $F=64$ learned filters with size $C=25$, ReLU activation, and a pooling window of 13.  The MICA feature responses were passed to the same MLP classifier discussed in Section \ref{section:spectral}, except it had a hidden-layer of 256 units to facilitate the higher dimensionality of the feature response.  MICA yielded a 4.8 percent increase in mean-class accuracy from the simpler MP experiment, demonstrating that unsupervised feature extraction can boost classification performance for high-resolution imagery.

The SCAE model used in this paper has the same architecture found in Figures \ref{fig:scae} and \ref{fig:scae_parts}.  It was trained with 30,000 $128\times 128$ image patches randomly extracted from the training and validation datasets.  Each CAE was trained individually with a batch size of 128.  Each convolution for the first convolution block has 32 units, 64 units for the second convolution block, 128 units for the third, and 256 units for the 1x1 convolution.  The refinement blocks have the same number of units as their corresponding convolution block, so the last hidden layer has 32 features.  

After training, the whole image is passed through the SCAE network to generate three $N\times 32$ feature responses.  These feature responses are concatenated, convolved with a $5 \times 5$ mean-pooling filter, and then reduced to 99\% of the original variance using WPCA.  The final feature response is passed to a MLP classifier with the same architecture used for the MICA model.  MICA outperformed SCAE showing that low-level features over a larger receptive field could be more important than higher-level features over a smaller spatial extent.

Table \ref{table:band_selection} shows the result of band-selection on classification performance.  The most significant impact to classification performance appears to be the NIR bands; consistent with the fact that most of the scene is vegetation. 

\begin{table}[ht!]
\centering
\setlength{\tabcolsep}{.25em}

\begin{tabular}{@{}lccc@{}}\toprule
& \textbf{OA} & \textbf{AA} & \textbf{$\kappa$} \\ \midrule
\textbf{SVM-RGB} & 36.4 & 19.9 & 0.274 \\
\textbf{SVM-NIR} & 59.6 & 26.0 & 0.493 \\
\textbf{SVM-CIR} & 50.2 & 25.3 & 0.401\\
\textbf{SVM-VNIR} & 52.4 & 25.7 & 0.444 \\
\textbf{SVM} & \textbf{61.3} & \textbf{29.6} & \textbf{0.538} \\
\bottomrule \\
\end{tabular}
\caption{The effect of band-selection on classification performance.  These results include per-class accuracy, overall accuracy (OA), mean-class accuracy (AA), and kappa statistic ($\kappa$).  For comparison, the last entry is the experiment from Table \ref{table:classification} which used all six-bands.}
\label{table:band_selection}
\end{table}

\subsection{Target Detection}
\label{section:targetdetection}
\begin{figure}[ht!]
\begin{center}
   \includegraphics[angle=90,width=0.9\linewidth]{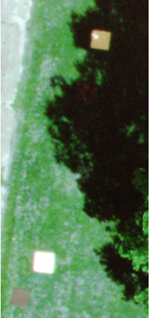}
\end{center}
   \caption{Both sets of black and white wooden targets.}
\label{fig:target}
\end{figure}

Our dataset also has a target detection challenge.  This challenge consists of two sets of white and black wooden panels, shown in Figure \ref{fig:target} where one set is placed in the shade and the other in direct sunlight.  The signature of both panels can be extracted using the training labels.    

This challenge is evaluated by the area under the curve metric.  Table \ref{table:targetdetection} has some benchmark results using commonly-used signature matched target detection algorithms, including the spectral angle mapper (SAM)~\cite{kruse1993spectral}, spectral-matched filter (SMF)~\cite{eismann2009automated}, constrained-energy minimization (CEM)~\cite{harsanyi1993detection}, and adaptive-cosine estimator (ACE)~\cite{kraut1999cfar}.  These benchmarks used global background estimations which were made by using every pixel in the image other than the targets.  These algorithms were performed both on the training and testing data, but no single algorithm worked  universally well on every target. 

\begin{table}[ht!]
\centering
\begin{tabular}{@{}lcccc@{}}\toprule
& \multicolumn{2}{c}{\textbf{Black Target}} & \multicolumn{2}{c}{\textbf{White Target}} \\
& \textbf{Train} & \textbf{Test} & \textbf{Train} & \textbf{Test} \\ \midrule
\textbf{SAM}            &\textbf{0.9716}&0.9188&0.8837&0.9054\\
\textbf{SMF (one-sided)}&0.9646&\textbf{0.9354}&0.9706&0.9258\\
\textbf{SMF (two-sided)}&0.6924&0.6130&\textbf{0.9798}&0.9318 \\
\textbf{CEM}            &0.3454&0.2927&0.9796&0.9284\\
\textbf{ACE}            &0.9614&0.9287&0.9125&\textbf{0.9331} \\
\bottomrule \\
\end{tabular}
\caption{Benchmark target detection results for the Hamlin Beach State Park dataset.  The results are the area under the curve detection metric for both black and white targets.  The detection results were calculated for both the training and testing set.}
\label{table:targetdetection}
\end{table}

\section{Discussion/Conclusion}

We have presented a dataset that will add to the growing repository of labeled non-RGB imagery.  We have separated the training, validation, and testing datasets to allow researchers to compare classification performance against current state-of-the-art performers.  We will work to make this data available on the IEEE GRSS evaluation server in order to standardize the evaluation of new semantic segmentation frameworks.

Our experimental results demonstrate the challenges associated with this dataset.  In addition, the large number of samples (pixels) present in this dataset carry a large-computational cost.  The MLP classifier and SCAE feature extraction framework were trained using the NVIDIA Titan X graphical processing unit (GPU) which has 12 GB of memory. None of the orthomosaics could be loaded into the GPU memory.  Future work for this dataset should involve an end-to-end classification framework that is capable of maximizing computer/GPU resources and making faster predictions than the approaches presented in this paper.

A classification framework that is capable of performing well on this dataset could be deployed in a real-time environment.  The separation of training/testing folds force the classification framework to make generalized predictions that can be transferred to other scenes.  A deployable model would also need to provide (near) real-time predictions, which is where an end-to-end deep-learning framework would do well.  The problem with this approach is that networks of this size would require an enormous labeled dataset collected with the Tetracam Micro-MCA6 sensor.

Future work will explore the development of an end-to-end DCNN framework inspired by current state-of-the-art semantic segmentation literature.  If a sufficient amount of training data is available, this method will be faster and likely more accurate.  

\section*{Acknowledgements}
We would like to thank Nina Raqueno, Paul Sponagle, Timothy Bausch, Michael McClelland II, and other members of the RIT Signature Interdisciplinary Research Area, UAS Research Laboratory that supported with data collection.

{\small
\bibliographystyle{ieee}
\bibliography{egbib}
}

\end{document}